\pdfoutput=1

\documentclass[11pt]{article}

\usepackage[final]{acl}

\usepackage{times}
\usepackage{latexsym}
\usepackage{booktabs}
\usepackage[T1]{fontenc}

\usepackage[utf8]{inputenc}

\usepackage{microtype}

\usepackage{inconsolata}

\usepackage{graphicx}
\usepackage{multirow}
\usepackage{makecell}
\usepackage{listings}
\usepackage{enumitem}
\usepackage{array}

\usepackage{CJKutf8}
\title{SRS-Stories: Using Large Language Models to Support Vocabulary Acquisition}
\title{SRS-Stories: Vocabulary-constrained multilingual story generation for language learning}

\author{Wiktor Kamzela$^1$ \and Mateusz Lango$^{1,2}$ \and Ond\v rej Du\v sek$^2$ \\
  $^1$Poznan University of Technology,  Institute of Computer Science, Poznan, Poland  \\
  $^2$Charles University, Faculty of Mathematics and Physics, Prague, Czechia \\
  \texttt{wiktor.kamzela@student.put.edu.pl},   \texttt{\{lango,odusek\}@ufal.mff.cuni.cz} \\}

\usepackage{adjustbox}
\usepackage{array}

\newcolumntype{R}[2]{%
    >{\adjustbox{angle=#1,lap=\width-(#2)}\bgroup}%
    l%
    <{\egroup}%
}
\newcommand*\rot{\multicolumn{1}{R{45}{1em}}}

\begin{document}
\maketitle
\begin{abstract}
In this paper, we use large language models to generate personalized stories for language learners, using only the vocabulary they know.
The generated texts are specifically written to teach the user new vocabulary by simply reading stories where it appears in context, while at the same time seamlessly reviewing recently learned vocabulary. The generated stories are enjoyable to read and the vocabulary reviewing/learning is optimized by a Spaced Repetition System.
The experiments are conducted in three languages: English, Chinese and Polish, evaluating three story generation methods and three strategies for enforcing lexical constraints. The results show that the generated stories are more grammatical, coherent, and provide better examples of word usage than texts generated by the standard constrained beam search approach.
\end{abstract}

\section{Introduction}
One of the very popular ways to learn a foreign language is to use a Spaced Repetition System (SRS)~\cite{leitner1972so,srs} such as Anki~\cite{elmes2015anki}, SuperMemo~\cite{wozniak1994optimization} or HackChinese.\footnote{\url{http://www.hackchinese.com}}
SRS is essentially an application for learning new vocabulary by reviewing a deck of flashcards. 
Every day, SRS selects a set of flashcards containing words learned in the past but predicted by the algorithm to be in danger of getting forgotten by the user in the near future. In this way, the system optimises the user's ability to recall all the vocabulary they have learned, while minimising the number of repetitions.
Although this strategy has proved effective in teaching the user to recall new words, it is of limited value for real vocabulary acquisition~\cite{aslan2011teaching,MILES2017103}. Words are reviewed out of context and learners often find it difficult to use them naturally in a sentence.  In addition, systematically reviewing flashcards with new vocabulary can become tedious.

Graded readers are resources that teach new vocabulary in context and are more interesting for the learner~\cite{nation1999graded}. 
These are books written with simplified vocabulary and with more difficult words accompanied by translations or definitions. The more difficult words are usually used several times in the story to help the reader remember them. While the advantages of this approach are clear, it may be difficult to find a graded reader at the appropriate language level (especially for languages other than English), and the learner has little control over the vocabulary studied. The repetition of learned words is not optimized in any way and learned vocabulary may be forgotten~\cite{waring2003rate}.

\begin{figure*}
    \centering
    \includegraphics[width=1\linewidth]{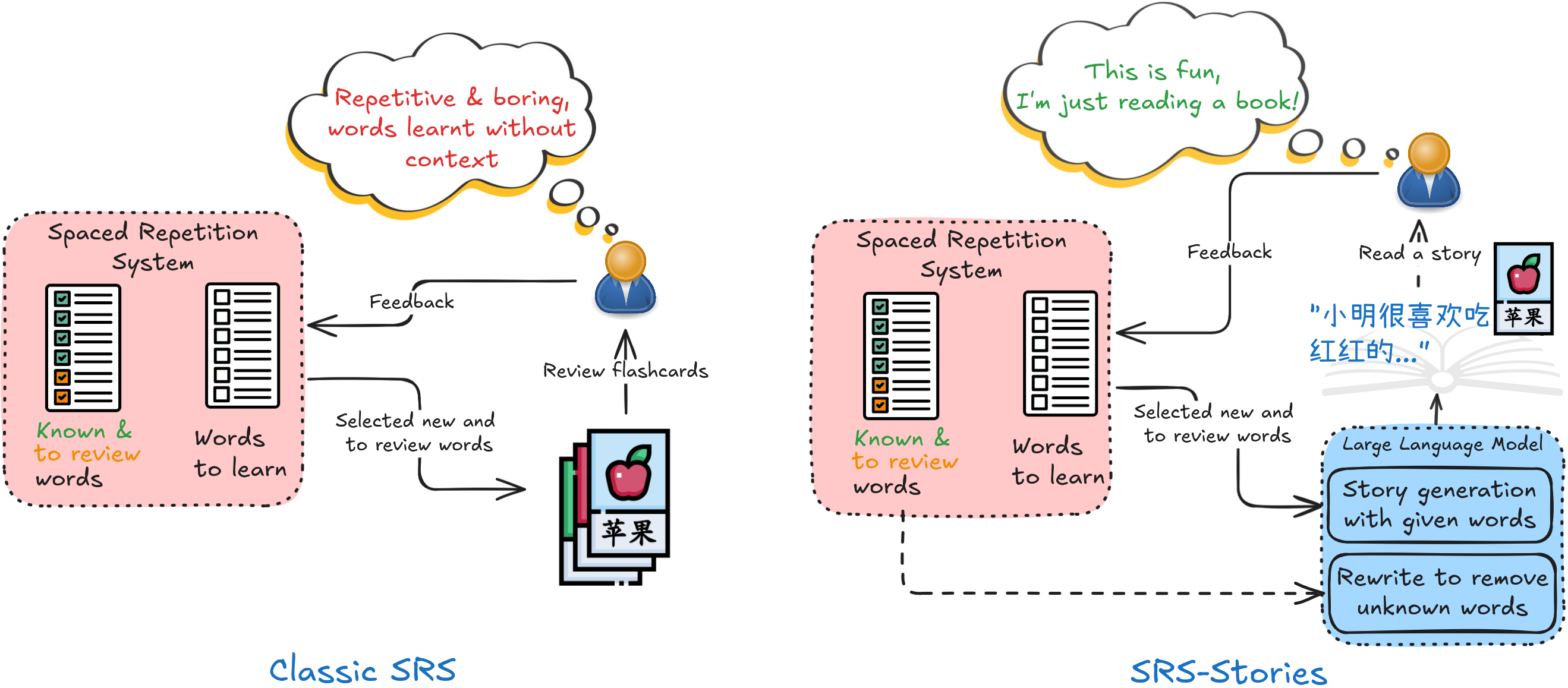}
    \caption{Overview of a classic SRS system and the presented approach: SRS-Stories.}
    \label{fig:overview}
\end{figure*}
In this paper, we explore the possibility of automatically generating graded readers (stories) that use vocabulary limited to words known to the user. In addition, the content of the story is specially designed to allow the introduction of selected new words, which are used several times in the story with meaningful context. The text can also deliberately include words that the user needs to review. Since such an approach can be coupled with a SRS system, effectively replacing flashcards with enjoyable stories, we call it SRS-Stories.

Our contributions are as follows:
\begin{itemize}[left=0em, itemsep=0em, topsep=0.2em]
\item We propose three prompting strategies for large language models to generate SRS-Stories,~i.e.~coherent texts written to teach words arbitrarily selected by SRS.
\item We design three strategies for enforcing vocabulary constraints based on text rewriting, involving an external non-neural constraint verifier and iterative story re-generation.
\item We carry out an experimental evaluation of the proposed story generation techniques at different language levels in three languages: English, Chinese and Polish. 
\end{itemize}
The results show that the proposed prompting strategies generate stories that are more interesting, coherent and grammatically correct than those generated by constrained beam search (CBS; \citealp{post-vilar-2018-fast,hu-etal-2019-improved}), a classic lexical constraint enforcement method.

\section{SRS-Stories}
SRS-Stories is a new language learning system that combines the benefits of graded readers with intelligent SRS vocabulary review schedules (see Fig.~\ref{fig:overview}
for an overview of the proposed approach). 
The system consists of a standard SRS engine and a large language model (LLM) with story generation capabilities.
The SRS engine stores the list of new words that the user wants to learn, as well as the list of previously learned words.
For each learning session, the SRS estimates which words need to be reviewed 
and picks a set of new words to learn.
Based on the selected words, the LLM generates a coherent story written with vocabulary appropriate to the learner.

The story contains the words selected by the SRS, allowing the user to review them seamlessly as they read.
In particular, new words are used multiple times in meaningful contexts in the generated text, naturally providing the user with examples of how the word can be used in sentences. 

\subsection{Task}   
The task considered in this paper is to generate a story under certain lexical constraints. Given a set of words $V$ known by the user and the words to learn $L$ selected by SRS, generate a coherent story $X=\{x_1, x_2, ..., x_n\}$ such that all words are suitable for the user ($x_i\in V \cup L$) and the new words appear at least $c$ times ($\forall_{l\in L} \sum_{x\in X}[\![x = l ]\!]\geq c$). In this paper we assume $c=3$.

\subsection{Methods}   

We propose three prompting strategies for generating a story that uses the selected words the required number of times.
These can be combined with three approaches for enforcing lexical constraints, which rewrite generated stories using only appropriate words ($ V \cup L$, i.e., previously known or selected new ones).
See Appx.~\ref{sec:prompts} for full prompts.

\subsubsection{Story generation strategies}
\paragraph{Simple Prompting}
We prompt the model to write a story of a certain length (in our experiments, 500-750 words) that uses selected words at least three times in sentences that clearly show their typical usage. 

\paragraph{Planning}
This method aims to generate more interesting and coherent stories with given words by dividing the story generation into three steps, allowing the model to better plan the generation: (1) the model is instructed to generate several ideas (\emph{titles}) for a story that can demonstrate the use of the given words. (2) the model is asked to select the most interesting idea and write an \emph{outline} for such a story. (3) the model is asked to turn the outline into a \emph{final story} of a given length, with each selected word used at least three times.

\paragraph{Examples First}
In this method, we first prompt the model to generate \emph{example sentences} for each selected word, before constructing the full story. This extra step is designed to help the model conceptualize the context and possible uses of the words, making their integration into the story more natural and coherent.

\subsubsection{Lexical constraint enforcement}
Once the story has been generated, it is processed by a non-LLM verifier. 
The verifier performs standard text normalization, tokenization, and lemmatization, then it checks the vocabulary used in the story against the list of words known to the user (see App.~\ref{app:verifier} for details). 
A list of broken lexical constraints, i.e.~words previously unknown to the user and not selected by SRS (i.e., $x_i \notin V \cup L$), is produced.

Then, we use the following three strategies to fix lexical constraints:

\paragraph{Rewrite (base)} 
We simply list the unknown words to the model and prompt it to replace them with simpler alternatives. This process is repeated five times.\footnote{In our experiments, the number of replaced words flattens out after 5 iterations, and subsequent iterations are less effective.} The full prompt is given in Appx.~\ref{sec:prompts}.

\paragraph{Rewrite Highlighted} 
In this strategy, instead of giving the LLM a list of words to replace, the model is asked to simplify words that are highlighted in the text. The story is presented to the model with unknown words surrounded by asterisks (Markdown-style bold text).

\paragraph{Get Synonyms then Rewrite} 
The model is first prompted to list synonyms for the listed words unknown to the user. In a second step, it is requested to rewrite the story using the synonyms provided.

\section{Experiments}
\subsection{Experimental setup} 
In order to evaluate\footnote{The implementation of our experiments is available \href{https://tinyurl.com/49sjs34u}{here}.} the proposed story generation techniques, we conduct experiments on three languages: English, Polish, and Chinese.

\paragraph{Vocabulary levels}
For each language, we compiled a vocabulary list with the assigned language proficiency level.
For English, we used the word lists provided by CEFR-J~\cite{negishi2013progress,tono2016cefrj}, a Japanese standard for English language proficiency.
For Chinese, the word lists from New HSK 3.0 standard were used~\cite{national2021chinese}.
Since no official word lists are available for Polish, we compiled them from a word frequency list computed on the Polish Wikipedia part of Leipzig Corpora Collection~\cite{goldhahn2012building}.
We assume the most frequent 1,500 words to be CEFR A1, 3,500 words for A2, 5,000 for B1, 7,500 for B2, 10,000 for C1, and all remaining words are assumed to be C2 words.

\paragraph{Study session setup}
We simulate a study session using the SRS-Stories system, assuming the user is familiar with all vocabulary at a given language level and aims to learn 10 random words from the next level.\footnote{We also experimented with a mixed setup, where some words are new (should have multiple occurrences in the story) and some are only reviewed (one occurrence is enough), but we found that teaching the user new words is much more challenging -- see Appx.~\ref{sec:mixedresults}.} 
Note that using random words is a challenging setup that does not exploit dependencies between words, such as being related to a single lesson topic.
For each language level and method, the results are averaged over 200 stories. 

\paragraph{LLMs used}
We generate the stories using the Llama 3.1 70B Instruct model~\cite{grattafiori2024llama}, an open-weight large language model that achieves high performance on many benchmarks and performed well in our preliminary experiments.
Additionally, we also report the performance of other LLMs for comparison. We selected: the general-purpose LLM
Qwen2.5 72B Instruct~\cite{qwen2025qwen25technicalreport}, the multilingual LLM Aya 23 35B~\cite{aryabumi2024aya23openweight} and the reasoning LLM DeepSeek R1 70B~\cite{deepseekai2025deepseekr1incentivizingreasoningcapability}.
The stories were generated using the default parameters in the vLLM library~\cite{kwon2023efficient}, except for the maximum number of tokens limited to 4096 and the temperature set to zero.

\paragraph{Baseline} 
As a baseline method of generating text with lexical constraints, we use the HuggingFace~\cite{wolf-etal-2020-transformers} implementation of Constrained Beam Search (CBS, \citealp{post-vilar-2018-fast,hu-etal-2019-improved}) to generate texts that specifically use given words $L$. 
Additionally, to limit the vocabulary only to words known by the user (i.e., use $V\cup L$ only), HuggingFace offers the “No Bad Words”  strategy of masking unwanted tokens at the output of the language model and setting their generation probability to zero. 
This approach adapts to multi-token words, masking out unwanted word continuations on-the-fly, based on tokens previously generated.

\subsection{Metrics}
\label{sec:metrics}
The stories are evaluated by a set of automatic metrics: 
(1) assessment of different aspects of the generated text by an LLM evaluator, (2) 
count-based metrics measuring the fulfillment of lexical constraints.
Additional metrics, including model-based ones, are provided in Appx.~\ref{app:detailed}.

\paragraph{LLM-based evaluation}
Previous works have shown great potential in using LLMs to imitate human evaluators in natural language generation (NLG) tasks~\cite{hu-etal-2024-themis,kocmi-federmann-2023-gemba}. 
Building upon this, we use an LLM to evaluate generated text. 
We prompt the model to return a score on a scale from 1 to 5 to assess: grammatical correctness (\textbf{Gram.}), text coherence (\textbf{Coh.}) and story interestingness (\textbf{Int.}). 
Following previous work~\cite{hu-etal-2024-themis,kartavc2025openlgauge}, we prompt the model to first list the errors found in the assessed text, and then provide a numerical quality assessment.
The exact prompts are provided in Appx.~\ref{app:evalprompts}.

As the evaluator, we use Qwen2.5 72B Instruct~\cite{qwen2025qwen25technicalreport}, which officially supports all languages used in our experiments. It also obtained the best results among open-weight models in a multilingual NLG evaluation presented by~\citet{chang2025exploringmultilingualnlgevaluation}.

\paragraph{Lexical constraints checks}
To evaluate the fulfillment of lexical constraints in generated stories, we compute several count-based metrics:
\begin{itemize}[left=0em, itemsep=0em, topsep=0.2em]
    \item \textbf{\# $L$} -- the average number of occurrences of each word to learn. The story should incorporate new vocabulary multiple times, providing various examples of word usage to reinforce learning and improve vocabulary retention.
    \item $\# |L| \geq 1$ -- the percentage of words to learn that are actually introduced in the story. 
    \item \textbf{Len.} -- the length of a generated story (number of  words for English and Polish, number of characters for Chinese).
    \item \textbf{OOV} -- the percentage of out-of-vocabulary/\hspace{0mm}unknown words ($x_i \notin V\cup L$). Ideally, the generated story should only use vocabulary that is known to the user. However, this is sometimes difficult to achieve, especially on lower language levels.
    
\end{itemize}

\begin{table*}[t]
 \centering \small
\begin{tabular}{llllllll}
   \toprule
\multicolumn{1}{l|}{} & \multicolumn{3}{l|}{LLM Scoring} & \multicolumn{2}{l|}{Average} & \multicolumn{2}{l}{Percent of} \\
\multicolumn{1}{l|}{Method} & Gram. & Coh. & \multicolumn{1}{l|}{Int.} & \# $L$ & \multicolumn{1}{l|}{Len.} & OOV & \multicolumn{1}{l}{\# $|L|\geq 1$} \\ \midrule
Constrained Beam Search & 3.22 & 3.82 & 3.92 & 1.46 & 538.9 & 6.71\% & 94.90\% \\\midrule
Simple prompting & \textbf{4.05} & 4.22 & 4.01 & \textbf{1.92} & 578.12 &  \textbf{0.64\%} & \textbf{95.40\%} \\
+ Get Synonyms then Rewrite & 3.92 & 4.21 & 4.12 & 1.91 & 599.85 & 5.03\% & 90.40\% \\
+ Rewrite Highlighted & 3.96 & \textbf{4.30} & \textbf{4.16} & 1.61 & 608.58 & 3.86\% & 81.10\% \\
\midrule
Examples first & \textbf{4.08} & \textbf{4.11} & 3.94 & \textbf{1.42} & 542.09 & \textbf{0.46\%} & \textbf{95.25\%} \\
+ Get Synonyms then Rewrite & 3.96 & 4.08 & \textbf{4.06} & 1.33 & 586.20 & 4.80\% & 92.15\% \\
 + Rewrite Highlighted & 3.94 & \textbf{4.11} & 3.99 & 1.21 & 576.83 & 4.08\% & 83.85\% \\
 \midrule
Planning & \textbf{3.98} & \textbf{4.17} & 4.04 & \textbf{2.68} & 609.67 & \textbf{0.01\%} & \textbf{96.35\%} \\
 + Get Synonyms then Rewrite & 3.85 & 4.16 & \textbf{4.16} & 2.45 & 645.57 & 0.05\% & 90.80\% \\
 + Rewrite Highlighted & 3.81 & 4.16 & 4.11 & 2.39 & 645.48 & 0.04\% & 83.75\% \\
\bottomrule
\end{tabular}
\caption{The results of different story generation strategies for English at B1 CEFR (lower intermediate) level. The metrics are defined in Sec.~\ref{sec:metrics}.}
\label{tab:eng}
\end{table*}

\begin{table*}[t]
 \centering \small
\begin{tabular}{llllllllll}
   \toprule
& \multicolumn{1}{l|}{} & \multicolumn{3}{l|}{LLM Scoring} & \multicolumn{2}{l|}{Average} & \multicolumn{2}{l}{Percent of} \\
Language & \multicolumn{1}{l|}{Method} & Gram. & Coh. & \multicolumn{1}{l|}{Int.} & \# $L$ & \multicolumn{1}{l|}{Len.} & OOV & \multicolumn{1}{l}{\# $|L|\geq 1$} \\ \midrule
\multirow{3}{*}{Polish} &
Simple prompting & 3.53 & 4.01 & 3.89 & \textbf{2.06} & 406.36 & \textbf{0.34\%} & \textbf{88.55\%} \\
& + Get Synonyms then Rewrite & 3.69 & 4.05 & \textbf{3.95} & 1.57 & 410.98 & 2.32\% & 65.85\% \\
& + Rewrite Highlighted & \textbf{3.81} & \textbf{4.08} & 3.92 & 1.52 & 408.36 & 3.45\% & 60.47\% \\ \midrule
\multirow{3}{*}{Chinese} &
Simple prompting & \textbf{3.99} & 4.06 & 3.61 & 0.68 & 804.41 & \textbf{1.72\%} & 18.93\% \\
& + Get Synonyms then Rewrite & 3.75 & \textbf{4.09} & \textbf{3.79} & 3.05 & 844.57 & 7.29\% & \textbf{79.63\%} \\
& + Rewrite Highlighted & 3.79 & \textbf{4.09} & 3.73 & \textbf{3.13} & 856.88 &  6.23\% & 78.24\% \\ \midrule
\multirow{3}{*}{English}  &
Simple prompting & \textbf{4.04} & 4.27 & 4.06 & \textbf{1.82} & 584.58 & \textbf{0.52\%} & \textbf{95.22\%} \\
& + Get Synonyms then Rewrite & 3.95 & 4.27 & \textbf{4.15} & 1.75 & 603.91 & 4.00\% & 88.73\% \\
& + Rewrite Highlighted & 3.97 & \textbf{4.31} & 4.14 & 1.18 & 618.64 & 3.11\% & 59.75\% \\
\bottomrule
\end{tabular}
\caption{The results of story generation for different languages, averaged over three language proficiency levels: lower intermediate (CEFR B1, HSK 3), upper intermediate (CEFR B2, HSK 4) and lower advanced (CEFR C1, HSK 5). The metrics are defined in Sec.~\ref{sec:metrics}.}
\label{tab:multilingual}
\end{table*}

\begin{table*}[t]
 \centering \small
\begin{tabular}{clllllllll}
   \toprule
\multicolumn{1}{l|}{} & \multicolumn{1}{l|}{} & \multicolumn{3}{l|}{LLM Scoring} & \multicolumn{2}{l|}{Average} & \multicolumn{2}{l}{Percent of} \\
\multicolumn{1}{l|}{Model} & \multicolumn{1}{l|}{Method} & Gram. & Coh. & \multicolumn{1}{l|}{Int.} & \# $L$ & \multicolumn{1}{l|}{Len.} & OOV & \multicolumn{1}{l}{\# $|L|\geq 1$} \\ \midrule
\multirow{3}{*}{\makecell{Llama 3.1 \\ 70B Instruct}} & Simple prompting & \textbf{4.04} & 4.27 & 4.06 & \textbf{1.82} & 584.58 & \textbf{0.52\%} & \textbf{95.22\%} \\
 & + Get Synonyms then Rewrite & 3.95 & 4.27 & \textbf{4.15} & 1.75 & 603.91 & 4.00\% & 88.73\% \\
 & + Rewrite Highlighted & 3.97 & \textbf{4.31} & 4.14 & 1.18 & 618.64 & 3.11\% & 59.75\% \\ \midrule
\multirow{3}{*}{\makecell{Qwen 2.5 \\ 72B Instruct}} & Simple prompting & \textbf{4.09} & \textbf{4.76} & 4.40 & 2.24 & 698.99 & \textbf{3.78\%} & 74.60\% \\
 & + Get Synonyms then Rewrite & 3.95 & 4.61 & \textbf{4.41} & \textbf{2.53} & 721.60 & 6.99\% & \textbf{83.55\%} \\
 & + Rewrite Highlighted & 4.03 & 4.65 & 4.39 & 1.40 & 720.89 & 6.29\% & 44.72\% \\ \midrule
\multirow{3}{*}{\makecell{DeepSeek R1 \\ 70B}} & Simple prompting & \textbf{4.03} & \textbf{4.57} & \textbf{4.42} & \textbf{1.83} & 692.78 & \textbf{5.52\%} & \textbf{77.72\%} \\
 & + Get Synonyms then Rewrite & 3.58 & 4.12 & 4.14 & 1.52 & 698.32 & 6.83\% & 65.07\% \\
 & + Rewrite Highlighted & 3.86 & 4.27 & 4.26 & 1.12 & 764.82 & 9.66\% & 46.58\% \\ \midrule
\multirow{3}{*}{\makecell{Aya 23 \\ 35B}} & Simple prompting & 4.05 & 4.37 & 4.02 & 0.60 & 314.06 & \textbf{6.42\%} & 33.52\% \\
 & + Get Synonyms then Rewrite & 4.08 & 4.70 & 4.39 & \textbf{1.49} & 652.74 & 8.92\% & \textbf{68.95\%} \\
 & + Rewrite Highlighted & \textbf{4.21} & \textbf{4.84} & \textbf{4.49} & 1.25 & 661.06 & 9.44\% & 52.82\% \\
\bottomrule
\end{tabular}
\caption{
The comparison of capabilities of different LLMs for English story generation,  averaged over B1, B2 and C1 CEFR levels.
The metrics are defined in Sec.~\ref{sec:metrics}.}
\label{tab:modelscomp}
\end{table*}

\section{Results}

\subsection{Story generation strategy comparison}
\label{sec:1stresults}

The results of the different story generation techniques are presented in Table~\ref{tab:eng} for English B1 (lower intermediate) language proficiency level.\footnote{Note that the story generation at the lower levels is more challenging due to the stronger lexical constraints.}  

CBS is the lowest-performing technique in terms of grammaticality, coherence and interestingness. 
This is confirmed by our manual analysis: stories generated by CBS sometimes list the new words out of context and the stories are not fluent, with frequent grammatical errors. 
They also often change the plot in completely unexpected ways, for example by suddenly changing the protagonist of the story.

Comparing the proposed story generation strategies, \emph{Simple prompting} achieved the highest coherence, the second-best grammaticality, and the highest interestingness -- ex aequo with \emph{Planning}.
The \emph{Planning} strategy used new words more frequently and produced slightly longer stories, though at the cost of worse grammatical accuracy.
The \emph{Example first} method yielded the highest grammatical correctness score but the lowest coherence and interestingness, exposing the learner least frequently to the target vocabulary.

With the base \emph{Rewrite} approach to removing unknown words, all story generation methods achieved the best grammatical correctness scores, the highest number of occurrences of the studied vocabulary $\#L$, and the lowest percentage of OOV words.
More advanced rewriting approaches led to higher interestingness scores across all story generation strategies and improved coherence in some variants, resulting in longer stories that, however, contained more words unfamiliar to the user.

\paragraph{Statistical analysis}
We also performed a statistical analysis of the English results, averaged across all language levels, using standard two-sample unpaired \textit{t}-tests. The detailed results are provided in App.~\ref{app:statanal}.

Across all three quality aspects measured by the LLM-based metric, the differences between the baseline CBS and any of the proposed strategies were statistically significant, with low $p$-values ($p < 0.001$).

For grammaticality, the differences between story generation methods were significant; however, the differences among the lexical constraint enforcement strategies were not statistically significant.

For coherence, the differences between stories produced by \emph{Simple prompting} under different rewriting strategies were again not significant, although any variant of \emph{Simple prompting} was statistically better than the other generation methods.
\emph{Examples first} was statistically significantly worse than \emph{Planning}.

Regarding interestingness, \emph{Simple prompting} with basic rewriting was significantly worse than when using more advanced rewriting strategies, but it was still significantly better than the other story generation methods.
Again, \emph{Examples first} was statistically significantly worse than \emph{Planning}.

\subsection{Multilingual evaluation}

Given the good results of \emph{Simple Prompting} 
and the higher computational cost of the other two  generation strategies (more iterations with LLM), we focused the multilingual experiments on \emph{Simple Prompting} only.
The results for Polish, Chinese and English, averaged over three language proficiency levels from lower intermediate to lower advanced, are shown in Table~\ref{tab:multilingual}.

The results for English averaged over different language levels are quite similar to the B1-level results in Sec.~\ref{sec:1stresults}, 
with high LLM-based scores and 
\emph{Rewrite} performing best at including new vocabulary into generated stories. 
The number of out-of-vocabulary words is lower, which is expected given that higher language levels use a wider base vocabulary.

The presented methods are also very successful in generating SRS-Stories in Chinese. The number of out-of-vocabulary words is relatively small and the new words are mentioned more than the required three times in the text.
The LLM assessment of grammatical correctness is the highest for \emph{Simple Prompting} with the simplest rewriting strategy (\emph{Rewrite}). 
However, \emph{Rewrite Highlighted} provides higher coherence, interestingness and is the most successful in frequently including the new words into the story. 
The high quality of the generated stories is also confirmed by our manual analysis, where we found the stories to be fluent and providing meaningful word usage examples in correct Chinese grammatical structures. 

For Polish, the stories generated by \emph{Simple Prompting} are shorter than for the other languages. \emph{Rewrite Highlighted} was the method that achieved the highest scores for grammaticality, coherence, and the second-best interestingness, but also a higher percentage of out-of-vocabulary words.
Simple \emph{Rewrite} was much more efficient in this respect, while also maximising the occurrence of new words.

\subsection{Results with different language models}

Table~\ref{tab:modelscomp} shows the comparison story generation capabilities of different LLMs. Among the investigated models, Llama 3.1 was the most successful in eliminating words unknown to the user from the generated text. It also was the most efficient in including all new vocabulary into the story. 

Qwen 2.5 achieved high scores on LLM-based metrics, but these may be biased since Qwen was also used as the evaluator~\cite{li2024llmsasjudgescomprehensivesurveyllmbased}. 
DeepSeek achieves the best results for \emph{Rewrite}, and using more advanced rewriting techniques seem to only harm the performance.  
Aya achieved the highest scores on grammatical correctness, coherence and interestingness, but was ineffective in including the new words into the story.

\subsection{Human evaluation}
\label{sec:human}
We a conducted human evaluation experiment, comparing our proposed \emph{Simple Prompting} strategy to the CBS baseline on 50 English stories generated by each method. To assess the multilingual capabilities of our approach, we also evaluated 50 Chinese and 50 Polish stories. The evaluation included five aspects on a 5-point Likert scale: grammatical correctness (\textbf{Gram.}), coherence (\textbf{Coh.}), interestingness (\textbf{Int.}), use of learned words in context (\textbf{Use}), and overall quality (\textbf{Over.}). See App.~\ref{app:guidelines} for annotation guidelines.

We recruited 20 annotators through the Prolific platform who had previously studied the story language as a second language and had reached fluency. This background ensures reliable judgments of grammatical correctness while also providing sensitivity to the language-learning context. Each annotator evaluated 10 stories, plus one additional story that served as an attention check (see App.~\ref{app:humevaldet}).

The results are presented in Tab.~\ref{tab:human}.
To compare our method with the baseline, we computed standard unpaired t-tests with 5\% significance level.
Whereas stories generated by CBS are of similar coherence and interestingness, the stories generated by the proposed approach are significantly more grammatically correct ($p=0.0151$) and significantly better illustrate the use of studied words ($p<0.0001$). Both these aspects are crucial in language learning applications.

Interestingly, the stories generated in Chinese are assessed by human annotators even better than those for English, achieving very high grammatical correctness, coherence, and scoring high on exemplary word usages.
The overall quality of Polish stories is assessed as higher than for English, with similar quality of word usage, coherence, and grammatical correctness.

\begin{table}[]
 \centering \small
\begin{tabular}{llllll}
   \toprule
Method & Gram.&Coh.&Int.&Use&Over.\\
\midrule 
English (CBS) & 3.56&3.96&3.23&1.24&3.24\\
English (Ours) & \textbf{4.08}&3.92&2.94&\textbf{3.88}&3.08\\
\midrule
Polish (Ours) & 3.78&3.92&3.50&	3.80&	3.66\\
\midrule
Chinese (Ours) & 4.68&4.40&	3.66&	4.40&	4.16\\

\bottomrule
\end{tabular}
\caption{The results of human evaluation of stories generated by Llama 3.1 70B with Constrained Beam Search (CBS) and  \emph{Simple Prompting} (Ours). 
Evaluation aspects are defined in Sec.~\ref{sec:human}. Statistically significant differences for English are in bold.
}
\label{tab:human}
\end{table}


\paragraph{Validation of LLM-based metrics}
\label{app:llmvalidation}
We run LLM-based evaluation metrics on the same stories that were assessed by human annotators and use the scores for Grammaticality, Coherence, and Interestingness to compute Pearson and Spearman correlations between the human and LLM judgments.

The results are presented in Table~\ref{tab:llmjudge}. 
The observed correlations with human judgments for English are moderate and comparable to those reported on story generation datasets for state-of-the-art LLM-based metrics~\cite{kartavc2025openlgauge}.
The correlations for Chinese and for Polish are lower, highlighting the challenges of using LLMs in multilingual contexts.

\begin{table}[]
    \centering\small \setlength{\tabcolsep}{5pt}
    \begin{tabular}{c|rrr|rrr}
    \toprule
    &\multicolumn{3}{c}{Pearson's $r$}&\multicolumn{3}{c}{Spearman's $\rho$}\\
         &  Gram.&Coh.&Int.&  Gram.&Coh.&Int.\\\midrule
        Eng.& 0.557&0.507&0.456  &0.383&0.280&0.465 \
        \\
        Ch.& 0.281&0.426&0.349  &0.298&0.326&0.336 \
        \\
        Pol.& 0.212&-0.149&0.070  &0.218&-0.158&0.092 \
        \\
        \bottomrule
    \end{tabular}
    \caption{Correlation between human annotators and LLM-based metrics.}
    \label{tab:llmjudge}
\end{table}

\section{Summary}
In this work, we present several methods for story generation for language learners with specific lexical constraints. 
The proposed story generation task can be coupled with a Spaced Repetition System (SRS) to teach the user new vocabulary via reading an enjoyable story, specially crafted for their language level and known vocabulary.

The experimental evaluation, conducted on three languages and various language levels,  demonstrated that modern LLMs prompted with the proposed strategies are capable to generate fluent, grammatically correct stories that introduce new vocabulary in context.
These results highlight the potential of SRS-Stories as emerging language learning technology.

\section*{Limitations}

The generated stories may reflect social biases present in the LLM’s pretraining data, influencing character roles or cultural assumptions. While the system aims to use new target words and avoid unknown vocabulary, some out-of-vocabulary (OOV) words may still appear, potentially affecting comprehension. Additionally, the narrative quality can vary, sometimes resulting in minor inconsistencies or unnatural phrasing. Future improvements could enhance bias mitigation and vocabulary control.

\section*{Acknowledgments}
This work was supported by the European Research Council (Grant agreement No.~101039303, NG-NLG) and used resources of the LINDAT/\hspace{0mm}CLARIAH-CZ Research Infrastructure (Czech Ministry of Education, Youth, and Sports project No. LM2018101).

\bibliography{custom,anthology}

\begin{thebibliography}{29}
\providecommand{\natexlab}[1]{#1}

\bibitem[{Aryabumi et~al.(2024)Aryabumi, Dang, Talupuru, Dash, Cairuz, Lin,
  Venkitesh, Smith, Campos, Tan, Marchisio, Bartolo, Ruder, Locatelli,
  Kreutzer, Frosst, Gomez, Blunsom, Fadaee, Üstün, and
  Hooker}]{aryabumi2024aya23openweight}
Viraat Aryabumi, John Dang, Dwarak Talupuru, Saurabh Dash, David Cairuz, Hangyu
  Lin, Bharat Venkitesh, Madeline Smith, Jon~Ander Campos, Yi~Chern Tan, Kelly
  Marchisio, Max Bartolo, Sebastian Ruder, Acyr Locatelli, Julia Kreutzer, Nick
  Frosst, Aidan Gomez, Phil Blunsom, Marzieh Fadaee, and 2 others. 2024.
\newblock \href {https://arxiv.org/abs/2405.15032} {Aya 23: Open weight
  releases to further multilingual progress}.
\newblock \emph{Preprint}, arXiv:2405.15032.

\bibitem[{Aslan(2011)}]{aslan2011teaching}
Yasin Aslan. 2011.
\newblock Teaching vocabulary effectively through flashcards.
\newblock \emph{International Journal of Arts \& Sciences}, 4(11):347.

\bibitem[{Bird and Loper(2004)}]{bird-loper-2004-nltk}
Steven Bird and Edward Loper. 2004.
\newblock \href {https://aclanthology.org/P04-3031/} {{NLTK}: The natural
  language toolkit}.
\newblock In \emph{Proceedings of the {ACL} Interactive Poster and
  Demonstration Sessions}, pages 214--217, Barcelona, Spain. Association for
  Computational Linguistics.

\bibitem[{Chang et~al.(2025)Chang, Gao, Hu, and
  Wan}]{chang2025exploringmultilingualnlgevaluation}
Jiayi Chang, Mingqi Gao, Xinyu Hu, and Xiaojun Wan. 2025.
\newblock \href {https://arxiv.org/abs/2503.04360} {Exploring the multilingual
  nlg evaluation abilities of llm-based evaluators}.
\newblock \emph{Preprint}, arXiv:2503.04360.

\bibitem[{DeepSeek-AI et~al.(2025)DeepSeek-AI, Guo, Yang, Zhang, Song, Zhang,
  Xu, Zhu, Ma, Wang, Bi, Zhang, Yu, Wu, Wu, Gou, Shao, Li, Gao, Liu, Xue, Wang,
  Wu, Feng, Lu, Zhao, Deng, Zhang, Ruan, Dai, Chen, Ji, Li, Lin, Dai, Luo, Hao,
  Chen, Li, Zhang, Bao, Xu, Wang, Ding, Xin, Gao, Qu, Li, Guo, Li, Wang, Chen,
  Yuan, Qiu, Li, Cai, Ni, Liang, Chen, Dong, Hu, Gao, Guan, Huang, Yu, Wang,
  Zhang, Zhao, Wang, Zhang, Xu, Xia, Zhang, Zhang, Tang, Li, Wang, Li, Tian,
  Huang, Zhang, Wang, Chen, Du, Ge, Zhang, Pan, Wang, Chen, Jin, Chen, Lu,
  Zhou, Chen, Ye, Wang, Yu, Zhou, Pan, Li, Zhou, Wu, Ye, Yun, Pei, Sun, Wang,
  Zeng, Zhao, Liu, Liang, Gao, Yu, Zhang, Xiao, An, Liu, Wang, Chen, Nie,
  Cheng, Liu, Xie, Liu, Yang, Li, Su, Lin, Li, Jin, Shen, Chen, Sun, Wang,
  Song, Zhou, Wang, Shan, Li, Wang, Wei, Zhang, Xu, Li, Zhao, Sun, Wang, Yu,
  Zhang, Shi, Xiong, He, Piao, Wang, Tan, Ma, Liu, Guo, Ou, Wang, Gong, Zou,
  He, Xiong, Luo, You, Liu, Zhou, Zhu, Xu, Huang, Li, Zheng, Zhu, Ma, Tang,
  Zha, Yan, Ren, Ren, Sha, Fu, Xu, Xie, Zhang, Hao, Ma, Yan, Wu, Gu, Zhu, Liu,
  Li, Xie, Song, Pan, Huang, Xu, Zhang, and
  Zhang}]{deepseekai2025deepseekr1incentivizingreasoningcapability}
DeepSeek-AI, Daya Guo, Dejian Yang, Haowei Zhang, Junxiao Song, Ruoyu Zhang,
  Runxin Xu, Qihao Zhu, Shirong Ma, Peiyi Wang, Xiao Bi, Xiaokang Zhang,
  Xingkai Yu, Yu~Wu, Z.~F. Wu, Zhibin Gou, Zhihong Shao, Zhuoshu Li, Ziyi Gao,
  and 181 others. 2025.
\newblock \href {https://arxiv.org/abs/2501.12948} {Deepseek-r1: Incentivizing
  reasoning capability in llms via reinforcement learning}.
\newblock \emph{Preprint}, arXiv:2501.12948.

\bibitem[{Devlin et~al.(2019)Devlin, Chang, Lee, and
  Toutanova}]{devlin-etal-2019-bert}
Jacob Devlin, Ming-Wei Chang, Kenton Lee, and Kristina Toutanova. 2019.
\newblock \href {https://doi.org/10.18653/v1/N19-1423} {{BERT}: Pre-training of
  deep bidirectional transformers for language understanding}.
\newblock In \emph{Proceedings of the 2019 Conference of the North {A}merican
  Chapter of the Association for Computational Linguistics: Human Language
  Technologies, Volume 1 (Long and Short Papers)}, pages 4171--4186,
  Minneapolis, Minnesota. Association for Computational Linguistics.

\bibitem[{Elmes(2015)}]{elmes2015anki}
D.~Elmes. 2015.
\newblock Anki.
\newblock \url{http://ankisrs.net}.

\bibitem[{Goldhahn et~al.(2012)Goldhahn, Eckart, Quasthoff
  et~al.}]{goldhahn2012building}
Dirk Goldhahn, Thomas Eckart, Uwe Quasthoff, and 1 others. 2012.
\newblock Building large monolingual dictionaries at the leipzig corpora
  collection: From 100 to 200 languages.
\newblock In \emph{LREC}, volume~29, pages 31--43.

\bibitem[{Grattafiori et~al.(2024)Grattafiori, Dubey, Jauhri, Pandey, Kadian,
  Al-Dahle, Letman, Mathur, Schelten, Vaughan et~al.}]{grattafiori2024llama}
Aaron Grattafiori, Abhimanyu Dubey, Abhinav Jauhri, Abhinav Pandey, Abhishek
  Kadian, Ahmad Al-Dahle, Aiesha Letman, Akhil Mathur, Alan Schelten, Alex
  Vaughan, and 1 others. 2024.
\newblock The llama 3 herd of models.
\newblock \emph{arXiv preprint arXiv:2407.21783}.

\bibitem[{Hu et~al.(2019)Hu, Khayrallah, Culkin, Xia, Chen, Post, and
  Van~Durme}]{hu-etal-2019-improved}
J.~Edward Hu, Huda Khayrallah, Ryan Culkin, Patrick Xia, Tongfei Chen, Matt
  Post, and Benjamin Van~Durme. 2019.
\newblock \href {https://doi.org/10.18653/v1/N19-1090} {Improved lexically
  constrained decoding for translation and monolingual rewriting}.
\newblock In \emph{Proceedings of the 2019 Conference of the North {A}merican
  Chapter of the Association for Computational Linguistics: Human Language
  Technologies, Volume 1 (Long and Short Papers)}, pages 839--850, Minneapolis,
  Minnesota. Association for Computational Linguistics.

\bibitem[{Hu et~al.(2024)Hu, Lin, Gao, Yin, and Wan}]{hu-etal-2024-themis}
Xinyu Hu, Li~Lin, Mingqi Gao, Xunjian Yin, and Xiaojun Wan. 2024.
\newblock \href {https://doi.org/10.18653/v1/2024.emnlp-main.891} {Themis: A
  reference-free {NLG} evaluation language model with flexibility and
  interpretability}.
\newblock In \emph{Proceedings of the 2024 Conference on Empirical Methods in
  Natural Language Processing}, pages 15924--15951, Miami, Florida, USA.
  Association for Computational Linguistics.

\bibitem[{Kart{\'a}{\v{c}} et~al.(2025)Kart{\'a}{\v{c}}, Lango, and
  Du{\v{s}}ek}]{kartavc2025openlgauge}
Ivan Kart{\'a}{\v{c}}, Mateusz Lango, and Ond{\v{r}}ej Du{\v{s}}ek. 2025.
\newblock {OpeNLGauge: An Explainable Metric for NLG Evaluation with
  Open-Weights LLMs}.
\newblock In \emph{Proceedings of the 18th International Natural Language
  Generation Conference}, Hanoi, Vietnam. Association for Computational
  Linguistics.

\bibitem[{Kieraś and Woliński(2017)}]{morfeusz}
Witold Kieraś and Marcin Woliński. 2017.
\newblock Morfeusz 2 – analizator i~generator fleksyjny dla języka
  polskiego.
\newblock \emph{Język Polski}, XCVII(1):75--83.

\bibitem[{Kocmi and Federmann(2023)}]{kocmi-federmann-2023-gemba}
Tom Kocmi and Christian Federmann. 2023.
\newblock \href {https://doi.org/10.18653/v1/2023.wmt-1.64} {{GEMBA}-{MQM}:
  Detecting translation quality error spans with {GPT}-4}.
\newblock In \emph{Proceedings of the Eighth Conference on Machine
  Translation}, pages 768--775, Singapore. Association for Computational
  Linguistics.

\bibitem[{Kwon et~al.(2023)Kwon, Li, Zhuang, Sheng, Zheng, Yu, Gonzalez, Zhang,
  and Stoica}]{kwon2023efficient}
Woosuk Kwon, Zhuohan Li, Siyuan Zhuang, Ying Sheng, Lianmin Zheng, Cody~Hao Yu,
  Joseph~E. Gonzalez, Hao Zhang, and Ion Stoica. 2023.
\newblock Efficient memory management for large language model serving with
  pagedattention.
\newblock In \emph{Proceedings of the ACM SIGOPS 29th Symposium on Operating
  Systems Principles}.

\bibitem[{Leitner(1972)}]{leitner1972so}
Sebastian Leitner. 1972.
\newblock So lernt man lernen: Der weg zum erfolg [learning to learn: The road
  to success].
\newblock \emph{Freiburg: Herder}.

\bibitem[{Li et~al.(2024)Li, Dong, Chen, Su, Zhou, Ai, Ye, and
  Liu}]{li2024llmsasjudgescomprehensivesurveyllmbased}
Haitao Li, Qian Dong, Junjie Chen, Huixue Su, Yujia Zhou, Qingyao Ai, Ziyi Ye,
  and Yiqun Liu. 2024.
\newblock \href {https://arxiv.org/abs/2412.05579} {Llms-as-judges: A
  comprehensive survey on llm-based evaluation methods}.
\newblock \emph{Preprint}, arXiv:2412.05579.

\bibitem[{Miles and Ehri(2017)}]{MILES2017103}
Katharine~Pace Miles and Linnea~C. Ehri. 2017.
\newblock \href {https://doi.org/10.1016/j.ecresq.2017.06.001} {Learning to
  read words on flashcards: Effects of sentence contexts and word class in
  native and nonnative english-speaking kindergartners}.
\newblock \emph{Early Childhood Research Quarterly}, 41:103--113.

\bibitem[{Nation and Wang(1999)}]{nation1999graded}
Paul Nation and Karen Wang. 1999.
\newblock Graded readers and vocabulary.

\bibitem[{{National Language Commission}(2021)}]{national2021chinese}
{National Language Commission}. 2021.
\newblock Chinese proficiency grading standards for international chinese
  language education.

\bibitem[{Negishi et~al.(2013)Negishi, Takada, and Tono}]{negishi2013progress}
Masashi Negishi, Tomoko Takada, and Yukio Tono. 2013.
\newblock A progress report on the development of the cefr-j.
\newblock In \emph{Exploring language frameworks: Proceedings of the ALTE
  Krak{\'o}w Conference}, pages 135--163.

\bibitem[{Post and Vilar(2018)}]{post-vilar-2018-fast}
Matt Post and David Vilar. 2018.
\newblock \href {https://doi.org/10.18653/v1/N18-1119} {Fast lexically
  constrained decoding with dynamic beam allocation for neural machine
  translation}.
\newblock In \emph{Proceedings of the 2018 Conference of the North {A}merican
  Chapter of the Association for Computational Linguistics: Human Language
  Technologies, Volume 1 (Long Papers)}, pages 1314--1324, New Orleans,
  Louisiana. Association for Computational Linguistics.

\bibitem[{Reddy et~al.(2016)Reddy, Labutov, Banerjee, and Joachims}]{srs}
Siddharth Reddy, Igor Labutov, Siddhartha Banerjee, and Thorsten Joachims.
  2016.
\newblock \href {https://doi.org/10.1145/2939672.2939850} {Unbounded human
  learning: Optimal scheduling for spaced repetition}.
\newblock In \emph{Proceedings of the 22nd ACM SIGKDD International Conference
  on Knowledge Discovery and Data Mining}, KDD '16, page 1815–1824, New York,
  NY, USA. Association for Computing Machinery.

\bibitem[{Tono(2016)}]{tono2016cefrj}
Yukio Tono. 2016.
\newblock \href {http://www.cefr-j.org/download.html} {The cefr-j wordlist
  version 1.6}.

\bibitem[{Waring and Takaki(2003)}]{waring2003rate}
Rob Waring and Misako Takaki. 2003.
\newblock At what rate do learners learn and retain new vocabulary from reading
  a graded reader?

\bibitem[{Warstadt et~al.(2019)Warstadt, Singh, and
  Bowman}]{warstadt-etal-2019-neural}
Alex Warstadt, Amanpreet Singh, and Samuel~R. Bowman. 2019.
\newblock \href {https://doi.org/10.1162/tacl_a_00290} {Neural network
  acceptability judgments}.
\newblock \emph{Transactions of the Association for Computational Linguistics},
  7:625--641.

\bibitem[{Wolf et~al.(2020)Wolf, Debut, Sanh, Chaumond, Delangue, Moi, Cistac,
  Rault, Louf, Funtowicz, Davison, Shleifer, von Platen, Ma, Jernite, Plu, Xu,
  Le~Scao, Gugger, Drame, Lhoest, and Rush}]{wolf-etal-2020-transformers}
Thomas Wolf, Lysandre Debut, Victor Sanh, Julien Chaumond, Clement Delangue,
  Anthony Moi, Pierric Cistac, Tim Rault, Remi Louf, Morgan Funtowicz, Joe
  Davison, Sam Shleifer, Patrick von Platen, Clara Ma, Yacine Jernite, Julien
  Plu, Canwen Xu, Teven Le~Scao, Sylvain Gugger, and 3 others. 2020.
\newblock \href {https://doi.org/10.18653/v1/2020.emnlp-demos.6} {Transformers:
  State-of-the-art natural language processing}.
\newblock In \emph{Proceedings of the 2020 Conference on Empirical Methods in
  Natural Language Processing: System Demonstrations}, pages 38--45, Online.
  Association for Computational Linguistics.

\bibitem[{Wo{\'z}niak and Gorzela{\'n}czyk(1994)}]{wozniak1994optimization}
Piotr Wo{\'z}niak and Edward Gorzela{\'n}czyk. 1994.
\newblock Optimization of repetition spacing in the practice of learning.
\newblock \emph{Acta neurobiologiae experimentalis}, 54(1):59--62.

\bibitem[{Yang et~al.(2025)Yang, Yang, Zhang, Hui, Zheng, Yu, Li, Liu, Huang,
  Wei, Lin, Yang, Tu, Zhang, Yang, Yang, Zhou, Lin, Dang, Lu, Bao, Yang, Yu,
  Li, Xue, Zhang, Zhu, Men, Lin, Li, Tang, Xia, Ren, Ren, Fan, Su, Zhang, Wan,
  Liu, Cui, Zhang, and Qiu}]{qwen2025qwen25technicalreport}
An~Yang, Baosong Yang, Beichen Zhang, Binyuan Hui, Bo~Zheng, Bowen Yu,
  Chengyuan Li, Dayiheng Liu, Fei Huang, Haoran Wei, Huan Lin, Jian Yang,
  Jianhong Tu, Jianwei Zhang, Jianxin Yang, Jiaxi Yang, Jingren Zhou, Junyang
  Lin, Kai Dang, and 23 others. 2025.
\newblock \href {https://arxiv.org/abs/2412.15115} {Qwen2.5 technical report}.
\newblock \emph{Preprint}, arXiv:2412.15115.

\end{thebibliography}

\appendix

\section{Frequently Asked Question}
\begin{enumerate}
    \item \emph{In the paper, $V$ is the set of words believed to be known by the user. However, the experiments are performed with the words assumed to be known according to CEFR/HSK lists which is different from the words really known to the user.}
    
While we conducted experiments in simulated environments where an artificial student was assumed to know all words at a given CEFR/HSK level, our system retrieves the list of known words directly from the SRS. The SRS records all previously learned vocabulary and dynamically separates it into two categories: words that require review (because they may have been forgotten or are close to being forgotten) and words that are retained (still known). Ideally, learners should use the SRS from the very beginning of their language learning journey.

If a learner does not start with the SRS and already possesses some vocabulary knowledge, most systems either estimate their vocabulary size basing on the language level (as in our experiments) or provide word lists aligned with textbooks to help the user construct the list of known words on their own. To bring the most benefits, SRS should store and schedule reviews of all known vocabulary. The challenge of initializing the SRS with a learner’s existing knowledge is not unique to our approach but is inherent to all SRS-based systems.

\item \emph{How are verb tenses handled by SRS-Stories?}

Lemmatization is applied for constraint verification i.e.~if the model generates a story containing different form of a required word, the system will accept it. Nevertheless, the model is asked to provide a story with a given verb form. Posing hard requirements on specific word forms would be impractical especially for Polish, which features relatively complex morphology not just for verbs but also nouns and other parts of speech.

\item \emph{How are phrases or idioms handled?}

Our experiments assume simple vocabulary for simplicity, but there is little difference in learning/using a lexicalized meaning of a longer phrase. Our approach has no technical limitations in this regard and can be directly used for phrase/idiom learning without any modifications.

\item \emph{How are function words handled in CEFR lists?}

Common function words such as "a" and "the" are part of the A1 English level. Other function words are often associated with specific tenses or grammatical structures (especially in Chinese) and are therefore assigned to different language levels according to their difficulty. The problem of learning how to use "a" or "the" is typically not related to recalling their meanings, so these words cannot be effectively learned using standard SRS. However, readers can gain an intuitive understanding of how to use these words based on the context by reading our SRS-Stories, which naturally use these function words frequently. This is one of the advantages of using our SRS-Stories vs. traditional SRS.

\end{enumerate}

\section{Prompts}
\label{sec:prompts}

Simple prompting 

\begin{lstlisting}
Write a story (500-750 words) containing words: [WORDS TO LEARN]. Use them in such a way that it is clear from the context of this story what they mean. Use each word at least three times. Write only story and nothing more.
\end{lstlisting}
Planning

\begin{lstlisting}
I would like to learn some new words in English. These words are: [WORDS TO LEARN]. Suggest some possible ideas for stories that can include all these words. Write only titles of stories and nothing more. Number stories with letters a, b, c... instead of numbers 1, 2, 3...
\end{lstlisting}

\begin{lstlisting}
Pick the most interesting idea. Write only the idea title and nothing more. Do not explain why you picked this one.
\end{lstlisting}

\begin{lstlisting}
Generate a plan in points for a story (500-750 words) based on the idea you have picked. It should be possible to generate a full story, based on this plan, which will contain all the words that I want to learn: [WORDS TO LEARN].
\end{lstlisting}

\begin{lstlisting}
Step by step generate a full story (500-750 words) based on this plan. Remember to use words which I want to learn: [WORDS TO LEARN]. Use them is such a way that it is clear from the context of this story what they mean. Use each word at least three times. Write only story and nothing more.
\end{lstlisting}
Examples first

\begin{lstlisting}
I will give you a list of words. For each of these words provide a sentence that will clearly illustrates what this word means without any prior knowledge of this word. Words for which you should generate sentences: [WORDS TO LEARN]
\end{lstlisting}

\begin{lstlisting}
Write a story (500-750 words) containing sentences that you have generated. If possible, try to use all of them. If not, remember to include all of the words [WORDS TO LEARN] in the story anyway. Make your story consistent and interesting to read. Write only a story and nothing more.
\end{lstlisting}
Rewrite

\begin{lstlisting}
Rewrite this story without these words: [UNKNOWN WORDS]. Use simpler alternatives of these words. Change only this words and nothing else. Do not remove these words: [WORDS TO LEARN]. Write only the story and nothing more. Do not inform me that here is a story.
\end{lstlisting}
Rewrite Highlighted

\begin{lstlisting}
I will provide you a story with some words marked with an asterisks (*) before and after a word, in other words these words are bolded. Change all of these words to simpler alternatives. Change only these words and nothing else. Do not remove these words: [WORDS TO LEARN]. Story with marked words: [HIGHLIGHTED STORY]. Write only the story and nothing more. Do not inform me that here is a story.
\end{lstlisting}
Get Synonyms then Rewrite

\begin{lstlisting}
In the following text there are several words taken in brackets (). Could you list for each of them a list of synonyms? [HIGHLIGHTED STORY]
\end{lstlisting}

\begin{lstlisting}
Re-write the above story by replacing the words in brackets with their synonyms. Don't do any other changes. Write only a story and nothing more.
\end{lstlisting}

\section{LLM-based evaluation prompts}
\label{app:evalprompts}

Grading grammar
\begin{lstlisting}
I will provide you a story. List all the grammatical errors in this story. Next give this story a grade from 1 to 5 (1 being the worst possible grade and 5 being the best possible grade). When grading a story focus on grammatical errors, do not take into account any other aspects. Story: [STORY]
\end{lstlisting}
Grading coherence

\begin{lstlisting}
I will provide you a story. Analyse whether this story is coherent or not. Next give this story a grade  from 1 to 5 (1 being the worst possible grade and 5 being the best possible grade). When grading a story, focus on its coherence, do not take into account grammatical errors.
\end{lstlisting}
Grading interestingness

\begin{lstlisting}
I will provide you a story. Analyse whether this story is interesting to read or not. Next give this story a grade from 1 to 5 (1 being the worst possible grade and 5 being the best possible grade). When grading a story focus on whether the story is interesting or not, do not take into account grammatical errors.
\end{lstlisting}

\section{Detailed results}
\label{app:detailed}
\label{sec:addresults}

Detailed results for English are in Table \ref{tab:englishall}, for Chinese in Table \ref{tab:chineseall}, and for Polish in Table \ref{tab:polishall}.

In addition to metrics shown in the main paper, we also report the following:
\begin{itemize}[left=0em, itemsep=0em, topsep=0.2em]
    \item \textbf{$L$ to len.} -- the ratio of new words to story length, provides an estimation of how often the new words occur in the story.
    \item Perplexity (\textbf{Perp.}) -- we use the Qwen2.5 7B model \cite{qwen2025qwen25technicalreport} to compute the perplexity of the generated text.
    \item We use the BERT model~\cite{devlin-etal-2019-bert} finetuned on Corpus of Linguistic Acceptability (\textbf{COLA}; \citealp{warstadt-etal-2019-neural}) to assess whether each sentence in the story is grammatically correct. We report the percentage of sentences assessed as correct.
    \item Next-sentence prediction (\textbf{NSP}) is a binary classification task that predicts whether two sentences appear consecutively in the text. We use the NSP head of the BERT model~\cite{devlin-etal-2019-bert} to assess the coherence of the story. 
\end{itemize}
Due to the suitability of models, the latter three metrics 
are computed for English only.

\begin{table*}[]
 \centering \small
 \setlength{\tabcolsep}{5pt}
\begin{tabular}{l>{\hspace{-2mm}}l|l>{\hspace{-2mm}}l>{\hspace{-1mm}}l|lll|l>{\hspace{-1mm}}l|lll}
   \toprule
 & \multicolumn{1}{l|}{} & \multicolumn{3}{l|}{LLM Scoring} & \multicolumn{3}{l|}{Average} & \multicolumn{2}{l|}{Percent of} & \multicolumn{3}{l}{Average}  \\
level & \multicolumn{1}{l|}{Method} & Gram. & Coh. & \multicolumn{1}{l|}{Int.} & \# $L$ & Len. & \hspace{-2mm}L to len. & OOV & \hspace{-2mm}\# $|L|\geq 1$ & NSP & Perp. & \hspace{-2mm}COLA \\
\midrule
B1 & Simple prompting & 4.05 & 4.22 & 4.01 & 1.92 & 578.12 & 29.69 & 0.64\% & 95.40\% & 0.96 & 5.16 & 0.98 \\
B1 & + Get Synonyms then R. & 3.92 & 4.21 & 4.12 & 1.91 & 599.85 & 30.86 & 5.03\% & 90.40\% & 0.96 & 5.55 & 0.98 \\
B1 & + Rewrite Highlighted & 3.96 & \textbf{4.30} & \textbf{4.16} & 1.61 & 608.58 & 37.17 & 3.86\% & 81.10\% & \textbf{0.97} & \textbf{5.01} & 0.98 \\
B1 & Planning & 3.98 & 4.17 & 4.04 & \textbf{2.68} & 609.67 & \textbf{22.34} & 0.52\% & \textbf{96.35\%} & 0.95 & 5.68 & 0.98 \\
B1 & + Get Synonyms then R. & 3.85 & 4.16 & \textbf{4.16} & 2.45 & 645.57 & 25.95 & 4.99\% & 90.80\% & 0.95 & 6.13 & 0.98 \\
B1 & + Rewrite Highlighted & 3.81 & 4.16 & 4.11 & 2.39 & 645.48 & 26.62 & 4.36\% & 83.75\% & 0.95 & 5.60 & 0.98 \\
B1 & Examples first & \textbf{4.08} & 4.11 & 3.94 & 1.42 & 542.09 & 37.66 & \textbf{0.46\%} & 95.25\% & 0.96 & 5.40 & \textbf{0.99} \\
B1 & + Get Synonyms then R. & 3.96 & 4.08 & 4.06 & 1.33 & 586.20 & 43.65 & 4.80\% & 92.15\% & 0.96 & 5.82 & \textbf{0.99} \\
B1 & + Rewrite Highlighted & 3.94 & 4.11 & 3.99 & 1.21 & 576.83 & 46.78 & 4.08\% & 83.85\% & 0.96 & 5.34 & 0.98 \\
\midrule
B2 & Simple prompting & 4.04 & \textbf{4.33} & 4.08 & 1.87 & 580.32 & 30.57 & \textbf{0.46\%} & 95.55\% & 0.96 & \textbf{4.91} & 0.98 \\
B2 & + Get Synonyms then R. & 3.96 & 4.31 & \textbf{4.16} & 1.70 & 605.04 & 35.44 & 3.64\% & 89.15\% & \textbf{0.97} & 5.15 & 0.98 \\
B2 & + Rewrite Highlighted & 4.00 & 4.32 & 4.12 & 1.22 & 600.93 & 48.84 & 2.59\% & 62.30\% & \textbf{0.97} & 4.93 & 0.98 \\
B2 & Planning & 3.99 & 4.18 & 4.07 & \textbf{2.33} & 616.34 & \textbf{26.39} & 0.56\% & 94.45\% & 0.95 & 5.47 & 0.98 \\
B2 & + Get Synonyms then R. & 3.90 & 4.19 & 4.11 & 2.05 & 643.70 & 31.08 & 3.40\% & 85.25\% & 0.96 & 5.63 & 0.98 \\
B2 & + Rewrite Highlighted & 3.90 & 4.16 & 4.07 & 1.79 & 647.58 & 35.86 & 3.30\% & 67.00\% & 0.95 & 5.30 & 0.98 \\
B2 & Examples first & \textbf{4.10} & 4.18 & 4.03 & 1.33 & 556.63 & 41.42 & 0.47\% & \textbf{95.80\%} & 0.96 & 5.14 & \textbf{0.99} \\
B2 & + Get Synonyms then R. & 3.98 & 4.17 & 4.10 & 1.26 & 588.54 & 46.53 & 3.54\% & 90.90\% & 0.96 & 5.25 & \textbf{0.99} \\
B2 & + Rewrite Highlighted & 3.95 & 4.20 & 4.03 & 0.97 & 589.09 & 59.56 & 3.03\% & 66.00\% & 0.96 & 4.99 & 0.98 \\
\midrule
C1 & Simple prompting & 4.04 & 4.27 & 4.09 & 1.68 & 595.29 & 35.27 & 0.46\% & 94.70\% & 0.96 & 5.11 & \textbf{0.98} \\
C1 & + Get Synonyms then R. & 3.98 & \textbf{4.29} & \textbf{4.16} & 1.63 & 606.83 & 36.96 & 3.34\% & 86.65\% & \textbf{0.97} & 5.16 & \textbf{0.98} \\
C1 & + Rewrite Highlighted & 3.94 & \textbf{4.29} & 4.13 & 0.70 & 646.41 & 92.02 & 2.87\% & 35.85\% & \textbf{0.97} & \textbf{4.92} & \textbf{0.98} \\
C1 & Planning & 3.97 & 4.23 & 4.05 & \textbf{2.10} & 624.02 & 29.57 & \textbf{0.45\%} & 93.70\% & 0.95 & 5.45 & \textbf{0.98} \\
C1 & + Get Synonyms then R. & 3.90 & 4.21 & 4.12 & 1.84 & 651.84 & 35.04 & 3.24\% & 82.35\% & 0.95 & 5.62 & \textbf{0.98} \\
C1 & + Rewrite Highlighted & 3.94 & 4.13 & 4.08 & 1.14 & 655.00 & 57.48 & 3.12\% & 45.55\% & 0.95 & 5.38 & \textbf{0.98} \\
C1 & Examples first & \textbf{4.08} & 4.16 & 4.04 & 1.21 & 574.54 & 47.19 & 0.46\% & \textbf{95.20\%} & 0.95 & 5.11 & \textbf{0.98} \\
C1 & + Get Synonyms then R. & 3.98 & 4.18 & 4.12 & 1.07 & 590.78 & 55.01 & 3.47\% & 87.05\% & 0.96 & 5.26 & \textbf{0.98} \\
C1 & + Rewrite Highlighted & 3.99 & 4.16 & 4.06 & 0.54 & 598.36 & \textbf{109.69} & 2.97\% & 42.80\% & 0.96 & 5.00 & \textbf{0.98} \\
\midrule
AVG & Simple prompting & 4.04 & 4.27 & 4.06 & 1.82 & 584.58 & 31.84 & 0.52\% & 95.22\% & 0.96 & 5.06 & 0.98 \\
AVG & + Get Synonyms then R. & 3.95 & 4.27 & \textbf{4.15} & 1.75 & 603.91 & 34.42 & 4.00\% & 88.73\% & \textbf{0.97} & 5.28 & 0.98 \\
AVG & + Rewrite Highlighted & 3.97 & \textbf{4.31} & 4.14 & 1.18 & 618.64 & 59.34 & 3.11\% & 59.75\% & \textbf{0.97} & \textbf{4.95} & 0.98 \\
AVG & Planning & 3.98 & 4.20 & 4.05 & \textbf{2.37} & 616.68 & \textbf{26.10} & 0.51\% & 94.83\% & 0.95 & 5.53 & 0.98 \\
AVG & + Get Synonyms then R. & 3.88 & 4.19 & 4.13 & 2.12 & 647.03 & 30.69 & 3.88\% & 86.13\% & 0.95 & 5.79 & 0.98 \\
AVG & + Rewrite Highlighted & 3.88 & 4.15 & 4.09 & 1.77 & 649.35 & 39.99 & 3.59\% & 65.43\% & 0.95 & 5.43 & 0.98 \\
AVG & Examples first & \textbf{4.09} & 4.15 & 4.00 & 1.32 & 557.75 & 42.09 & \textbf{0.46\%} & \textbf{95.42\%} & 0.95 & 5.22 & \textbf{0.99} \\
AVG & + Get Synonyms then R. & 3.97 & 4.14 & 4.09 & 1.22 & 588.51 & 48.39 & 3.94\% & 90.03\% & 0.96 & 5.44 & \textbf{0.99} \\
AVG & + Rewrite Highlighted & 3.96 & 4.16 & 4.03 & 0.91 & 588.09 & 72.01 & 3.36\% & 64.22\% & 0.96 & 5.11 & 0.98 \\

\bottomrule
\end{tabular}
\caption {Detailed results for English, Llama-3.1-70B-Instruct}
\label{tab:englishall}
\end{table*}

\begin{table*}[]
 \centering \small
\begin{tabular}{ll|lll|lll|ll}
\toprule
 & \multicolumn{1}{l|}{} & \multicolumn{3}{l|}{LLM Scoring} & \multicolumn{3}{l|}{Average} & \multicolumn{2}{l}{Percent of} \\
level & \multicolumn{1}{l|}{method} & Gram. & Coh. & \multicolumn{1}{l|}{Int.} & \# $L$ & Len. & \multicolumn{1}{l|}{$L$ to len.} & OOV & \multicolumn{1}{l}{\# $|L|\geq 1$} \\
\midrule
HSK 3 & Simple prompting & \textbf{3.94} & 4.09 & 3.54 & 0.94 & 809.93 & 86.55 & 2.77\% & 27.38\% \\
HSK 3 & + Get Synonyms then Rewrite & 3.63 & \textbf{4.15} & \textbf{3.77} & 3.13 & 849.03 & 26.65 & \textbf{10.11\%} & \textbf{81.38\%} \\
HSK 3 & + Rewrite Highlighted & 3.68 & 4.09 & 3.70 & \textbf{3.43} & 881.21 & \textbf{24.78} & 8.23\% & 81.26\% \\
\midrule
HSK 4 & Simple prompting & \textbf{3.98} & 4.08 & 3.65 & 0.77 & 804.31 & \textbf{104.57} & \textbf{1.73\%} & 21.49\% \\
HSK 4 & + Get Synonyms then Rewrite & 3.71 & 4.05 & \textbf{3.78} & 2.60 & 840.04 & 32.33 & 5.83\% & 77.41\% \\
HSK 4 & + Rewrite Highlighted & 3.79 & \textbf{4.09} & 3.70 & \textbf{2.87} & 846.47 & 28.84 & 5.34\% & \textbf{78.96\%} \\
\midrule
HSK 5 & Simple prompting & \textbf{4.05} & 4.01 & 3.65 & 0.33 & 798.98 & 244.53 & \textbf{0.65\%} & 7.91\% \\
HSK 5 & + Get Synonyms then Rewrite & 3.92 & \textbf{4.08} & \textbf{3.83} & \textbf{3.41} & 844.63 & \textbf{24.74} & 5.94\% & \textbf{80.10\%} \\
HSK 5 & + Rewrite Highlighted & 3.88 & \textbf{4.08} & 3.78 & 3.10 & 842.95 & 26.74 & 5.11\% & 74.50\% \\
\midrule
AVG & Simple prompting & \textbf{3.99} & 4.06 & 3.61 & 0.68 & 804.41 & \textbf{145.22} & \textbf{1.72\%} & 18.93\% \\
AVG & + Get Synonyms then Rewrite & 3.75 & \textbf{4.09} & \textbf{3.79} & 3.05 & 844.57 & 27.90 & 7.29\% & \textbf{79.63\%} \\
AVG & + Rewrite Highlighted & 3.79 & \textbf{4.09} & 3.73 & \textbf{3.13} & 856.88 & 26.79 & 6.23\% & 78.24\% \\

\bottomrule
\end{tabular}
\caption {Detailed results for Chinese, Llama-3.1-70B-Instruct}
\label{tab:chineseall}
\end{table*}

\begin{table*}[]
 \centering \small
\begin{tabular}{ll|lll|lll|ll}
\toprule
 & \multicolumn{1}{l|}{} & \multicolumn{3}{l|}{LLM Scoring} & \multicolumn{3}{l|}{Average} & \multicolumn{2}{l}{Percent of} \\
level & \multicolumn{1}{l|}{method} & Gram. & Coh. & \multicolumn{1}{l|}{Int.} & \# $L$ & Len. & \multicolumn{1}{l|}{$L$ to len.} & OOV & \multicolumn{1}{l}{\# $|L|\geq 1$} \\
\midrule
B1 & Simple prompting & 3.54 & 4.08 & 3.89 & \textbf{2.04} & 390.99 & \textbf{19.14} & \textbf{0.48\%} & \textbf{89.60\%} \\
B1 & + Get Synonyms then Rewrite & 3.69 & \textbf{4.13} & \textbf{3.96} & 1.57 & 388.01 & 24.69 & 3.31\% & 69.45\% \\
B1 & + Rewrite Highlighted & \textbf{3.78} & 4.11 & 3.93 & 1.61 & 386.63 & 24.07 & 5.20\% & 69.50\% \\
\midrule
B2 & Simple prompting & 3.59 & 4.03 & 3.89 & \textbf{1.95} & 402.29 & \textbf{20.59} & \textbf{0.22\%} & \textbf{87.80\%} \\
B2 & + Get Synonyms then Rewrite & 3.64 & 4.07 & \textbf{3.98} & 1.50 & 396.50 & 26.45 & 1.91\% & 69.80\% \\
B2 & + Rewrite Highlighted & \textbf{3.83} & \textbf{4.11} & 3.93 & 1.46 & 403.66 & 27.74 & 2.68\% & 64.85\% \\
\midrule
C1 & Simple prompting & 3.46 & 3.94 & 3.90 & \textbf{2.17} & 425.81 & \textbf{19.63} & \textbf{0.33\%} & \textbf{88.25\%} \\
C1 & + Get Synonyms then Rewrite & 3.75 & 3.95 & \textbf{3.92} & 1.64 & 448.43 & 26.50 & 1.73\% & 58.30\% \\
C1 & + Rewrite Highlighted & \textbf{3.83} & \textbf{4.02} & 3.88 & 1.49 & 434.78 & 28.09 & 2.47\% & 47.05\% \\
\midrule
AVG & Simple prompting & 3.53 & 4.01 & 3.89 & \textbf{2.06} & 406.36 & \textbf{19.79} & \textbf{0.34\%} & \textbf{88.55\%} \\
AVG & + Get Synonyms then Rewrite & 3.69 & 4.05 & \textbf{3.95} & 1.57 & 410.98 & 25.88 & 2.32\% & 65.85\% \\
AVG & + Rewrite Highlighted & \textbf{3.81} & \textbf{4.08} & 3.92 & 1.52 & 408.36 & 26.63 & 3.45\% & 60.47\% \\

\bottomrule
\end{tabular}
\caption {Detailed results for Polish, Llama-3.1-70B-Instruct}
\label{tab:polishall}
\end{table*}

\begin{table*}[]

 \centering \small
 \setlength{\tabcolsep}{5pt}
\begin{tabular}{ll|lll|ll|ll|ll|lll}
   \toprule
      &                  &       &      & \multicolumn{1}{l|}{} & \multicolumn{2}{l|}{New+recall}      & \multicolumn{2}{l|}{New}             & \multicolumn{2}{l|}{Recall}          &         &        \\
      &                  & \multicolumn{3}{l|}{LLM Scoring}     & Avg. & \multicolumn{1}{l|}{\% of} & Avg. & \multicolumn{1}{l|}{\% of} & Avg. & \multicolumn{1}{l|}{\% of} & Avg. & \% of  \\
level & method           & Gram. & Coh. & Int.                  & \# $L$  & \# $|L|\geq 1$             & \# $L$  & \# $|L|\geq 1$             & \# $L$  & \# $|L|\geq 1$             & Len.    & OOV    \\
\midrule
B1     & Simple pr. & 4.04  & 4.34 & 4.00                  & 4.47    & 93.20\%                    & 3.33    & 29.60\%                    & 1.38    & 63.60\%                    & 572.76  & 4.39\% \\
B2     & Simple pr. & 3.94  & 4.22 & 4.04                  & 2.87    & 94.40\%                    & 3.48    & 29.60\%                    & 1.19    & 64.80\%                    & 548.54  & 2.93\% \\
C1     & Simple pr. & 4.02  & 4.32 & 4.14                  & 3.31    & 91.40\%                    & 3.05    & 28.40\%                    & 1.17    & 63.00\%                    & 563.48  & 3.04\% \\
\midrule
AVG     & Simple pr. & 4.00  & 4.29 & 4.06                  & 3.55    & 93.00\%                    & 3.29    & 29.20\%                    & 1.25    & 63.80\%                    & 561.59  & 3.45\% \\      
\bottomrule
\end{tabular}
\caption {English, Llama-3.1-70B-Instruct, Words to be learned divided into new and recall sets}
\label{tab:mixedsetup}
\end{table*}

\section{Results for mixed setup}
\label{sec:mixedresults}
We also conducted preliminary experiments with Simple Prompting in a setup where the story would not only contain new words that would be used multiple times in meaningful contexts, but also review words that could only be mentioned.
In our experimental setup, we used 3 random words as new words and 7 random words as words to review.
The results are presented in Tab.~\ref{tab:mixedsetup}.

\section{Details on lexical constraints verification}
\label{app:verifier}
To check the lexical constraints, the Polish text was preprocessed using simple tokenisation with whitespaces and lemmatisation with  Morfeusz~\cite{morfeusz}. Next, all possible conjugations/declinations of the lemmas were extracted from Morfeusz and checked against the list of permitted vocabulary.
For the English stories, tokenisation and lemmatisation were performed using the nltk library (\texttt{WordNetLemmatizer})~\cite{bird-loper-2004-nltk}.  No tokenisation or lemmatisation was applied to Chinese. 
Therefore, to assess out-of-vocabulary metrics, we counted the number of Chinese characters that could not be mapped to the vocabulary lists for the given level.
Before processing the text, all punctuation was removed and the text was converted to lowercase for all languages.

\section{Human evaluation details}
\label{app:humevaldet}
The task of annotating eleven stories was estimated to take 45 minutes. The annotators were paid £6 for task completion (equivalent to £8 per hour), which is within the 'fair' pay range on the Prolific platform. One of the stories was used as an attention check: the order of the paragraphs had been switched (coherence), many obvious grammatical errors had been introduced manually (correctness) and none of the studied words were used (word use). Annotators who assessed these aspects as higher than 3 (the middle of the score range) were rejected from the study. The average assessment of the story used for the attention check is provided in Table~\ref{tab:fakestory}.

We used four screening filters provided by Prolific for annotators selection:
1) Annotators had to be fluent in the language of the story (English, Polish or Chinese).
2) The language of the story had to be a language other than the annotators' first language.
3) The task approval rate had to be above 95\%.
4) The number of tasks performed on the platform had to be above 50. 
The annotation was performed via Google Forms.
The study was approved by the Ethics Committee of the authors' institution.

\begin{table}[]
 \centering \small
\begin{tabular}{llllll}
   \toprule
Method & Gram.&Coh.&Int.&Use&Over.\\
\midrule 
English  & 1.90&1.90&1.70&1.10&1.30\\
Polish  & 2.60&	3.60&3.20&	1.00&	2.80\\
Chinese & 2.00&2.20&	1.40&	1.00&	1.80\\
\bottomrule
\end{tabular}
\caption{The results of human evaluation of the story with artificially introduced errors, which served as an attention check for the annotators.
}
\label{tab:fakestory}
\end{table}




\section{Statistical analysis}
\label{app:statanal}
The results of two-sample unpaired T-test performed for English stories are presented in Tables~\ref{tab:stat-en-gram}, \ref{tab:stat-en-coh}, and \ref{tab:stat-en-int}.
\begin{table*}[]
\centering\small \setlength{\tabcolsep}{5pt}
\begin{tabular}{llll|lll|lll}
\toprule
              &\rot{Simple p.} &\rot{+ Get Syn.} &\rot{+ Rewrite.} &\rot{Example first} &\rot{+ Get Syn.} &\rot{+ Rewrite.} &\rot{Planning} &\rot{+ Get Syn.} &\rot{+ Rewrite.} \\\midrule
CBS           & 0.000     & 0.000      & 0.000      & 0.000         & 0.000      & 0.000      & 0.000    & 0.000      & 0.000      \\
Simple prompting     & -         & 0.000      & 0.000      & 0.009         & 0.000      & 0.000      & 0.001    & 0.000      & 0.000      \\
+ Get Synonyms then Rewrite     & -         & -          & 0.153      & 0.000         & 0.125      & 0.278      & 0.037    & 0.000      & 0.000      \\
+ Rewrite Highlighted    & -         & -          & -          & 0.000         & 0.460      & 0.352      & 0.220    & 0.000      & 0.000      \\
Example first & -         & -          & -          & -             & 0.000      & 0.000      & 0.000    & 0.000      & 0.000      \\
+ Get Synonyms then Rewrite     & -         & -          & -          & -             & -          & 0.314      & 0.246    & 0.000      & 0.000      \\
+ Rewrite Highlighted    & -         & -          & -          & -             & -          & -          & 0.133    & 0.000      & 0.000      \\
Planning      & -         & -          & -          & -             & -          & -          & -        & 0.000      & 0.000      \\
+ Get Synonyms then Rewrite   & -         & -          & -          & -             & -          & -          & -        & -          & 0.467      \\
+ Rewrite Highlighted    & -         & -          & -          & -             & -          & -          & -        & -          & -       \\\bottomrule  
\end{tabular}
\caption{P-values of two-sample unpaired T-tests between the results of \emph{Grammaticality} obtained by different generation methods for English, averaged across all levels. }
\label{tab:stat-en-gram}
\end{table*}

\begin{table*}[]
\centering\small \setlength{\tabcolsep}{5pt}
\begin{tabular}{llll|lll|lll}
\toprule
              &\rot{Simple p.} &\rot{+ Get Syn.} &\rot{+ Rewrite.} &\rot{Example first} &\rot{+ Get Syn.} &\rot{+ Rewrite.} &\rot{Planning} &\rot{+ Get Syn.} &\rot{+ Rewrite.} \\\midrule
CBS           & 0.000     & 0.000      & 0.000      & 0.000         & 0.000      & 0.000      & 0.000    & 0.000      & 0.000      \\
Simple prompting     & -         & 0.424      & 0.126      & 0.000         & 0.000      & 0.000      & 0.001    & 0.000      & 0.000      \\
+ Get Synonyms then Rewrite    & -         & -          & 0.092      & 0.000         & 0.000      & 0.000      & 0.001    & 0.000      & 0.000      \\
+ Rewrite Highlighted    & -         & -          & -          & 0.000         & 0.000      & 0.000      & 0.000    & 0.000      & 0.000      \\
Example first & -         & -          & -          & -             & 0.350      & 0.408      & 0.022    & 0.051      & 0.469      \\
+ Get Synonyms then Rewrite    & -         & -          & -          & -             & -          & 0.266      & 0.008    & 0.021      & 0.376      \\
+ Rewrite Highlighted    & -         & -          & -          & -             & -          & -          & 0.035    & 0.076      & 0.376      \\
Planning      & -         & -          & -          & -             & -          & -          & -        & 0.358      & 0.016      \\
+ Get Synonyms then Rewrite    & -         & -          & -          & -             & -          & -          & -        & -          & 0.040      \\
+ Rewrite Highlighted    & -         & -          & -          & -             & -          & -          & -        & -          & -         \\\bottomrule
\end{tabular}
\caption{P-values of two-sample unpaired T-tests between the results of \emph{Coherence} obtained by different generation methods for English, averaged across all levels. }\label{tab:stat-en-coh}
\end{table*}

\begin{table*}[]
\centering\small \setlength{\tabcolsep}{5pt}
\begin{tabular}{llll|lll|lll}
\toprule
              &\rot{Simple p.} &\rot{+ Get Syn.} &\rot{+ Rewrite.} &\rot{Example first} &\rot{+ Get Syn.} &\rot{+ Rewrite.} &\rot{Planning} &\rot{+ Get Syn.} &\rot{+ Rewrite.} \\\midrule
CBS & 0.000 & 0.000 & 0.000 & 0.000 & 0.000 & 0.000 & 0.000 & 0.000 & 0.000\\
Simple prompting & - & 0.000 & 0.000 & 0.000 & 0.061 & 0.018 & 0.255 & 0.000 & 0.068\\
+ Get Synonyms then Rewrite & - & - & 0.412 & 0.000 & 0.005 & 0.000 & 0.000 & 0.205 & 0.001\\
+ Rewrite Highlighted & - & - & - & 0.000 & 0.013 & 0.000 & 0.000 & 0.294 & 0.006\\
Example first & - & - & - & - & 0.000 & 0.083 & 0.001 & 0.000 & 0.000\\
+ Get Synonyms then Rewrite & - & - & - & - & - & 0.000 & 0.011 & 0.030 & 0.428\\
+ Rewrite Highlighted & - & - & - & - & - & - & 0.047 & 0.000 & 0.000\\
Planning & - & - & - & - & - & - & - & 0.000 & 0.010\\
+ Get Synonyms then Rewrite & - & - & - & - & - & - & - & - & 0.013\\
+ Rewrite Highlighted & - & - & - & - & - & - & - & - & -
\\\bottomrule
\end{tabular}
\caption{P-values of two-sample unpaired T-tests between the results of \emph{Interestingness} obtained by different generation methods for English, averaged across all levels. }
\label{tab:stat-en-int}
\end{table*}

\section{Annotation guidelines}
\label{app:guidelines}
This task concerns the evaluation of short stories (texts) intended as learning material for students of Chinese at different levels. Each story is accompanied by a list of words that can be learned by reading it.

Your task is to carefully read 11 such stories and their word lists in order to answer the following questions:
\begin{itemize}[left=0em, itemsep=0em, topsep=0.2em]
    \item How fluent (i.e. grammatically correct) is the story?
\item  How coherent is the story (i.e. is the plot logical and consistent)? 
\item  How interesting/enjoyable to read is the story (given the language learning context)?
\item  How well does the story illustrate the use of words from the list ? Are the words from the word list used in the story, and are they used correctly?  Does their use help to guess their meaning? 
\item  What is your overall evaluation of the generated story?
\end{itemize}
You should answer each question by selecting a score from 1 (lowest) to 5 (highest).

Warning: The stories are generated by an Artificial Intelligence algorithm and are entirely fictional.

\section{Examples of generated stories}
\label{app:examples}
\subsection{English}
An example of a story generated at B2 level for words \emph{carriage, core, environmentally, imagine, relaxing, reluctant, sure, timely, tuition, unexpectedly} is provided below.

In the small town, in the heart of the countryside, a beautiful old \emph{carriage} stood proudly in the town square. The \emph{carriage}, with its carefully carved wooden panels and soft seats, had been a part of the town's history for generations. It was said that the \emph{carriage} had once belonged to a wealthy family who had used it to travel to and from their grand estate on the outskirts of town.

As a child, she had always been fascinated by the \emph{carriage}. She would often think about herself riding in it, feeling like a princess on her way to a grand ball. But as she grew up, her fascination with the \emph{carriage} never went away. In fact, it became stronger. She began to see the \emph{carriage} as a symbol of the town's rich history and a reminder of the importance of preserving the past at its \emph{core}.

As she entered her adult years, she became increasingly passionate about living in a way that helped the \emph{environment}. She believed that it was essential to live in a way that was good for the planet and to reduce the harm we cause. She began to think of ways to restore the old \emph{carriage} in a way that would be good for the \emph{environment}. She wanted to make \emph{sure} that the \emph{carriage} would be a place where people could go to \emph{relax} and enjoy the peaceful atmosphere of the countryside.

Her friends and family were initially uncertain to support her idea. They thought it would be a waste of time and money to restore the old \emph{carriage}. But she was certain that it was a project worth doing. She spent many hours researching and planning, determined to find a way to make it work in a \emph{timely} manner.

Finally, after months of hard work, her vision began to take shape. She had found a team of skilled workers who were willing to work with her to restore the \emph{carriage} using methods that were good for the \emph{environment}. The project was not without its challenges, but she was determined to see it through.

As the work progressed, the people of the town began to take notice. They would stop and watch as she and her team worked hard to bring the \emph{carriage} back to life. Slowly but \emph{surely}, the \emph{carriage} began to transform before their eyes. The wooden panels were sanded and polished, the seats were made new again, and the metal parts were restored to their former shine.

When the work was complete, the people of the town gathered to admire the finished product. The \emph{carriage} was more beautiful than anyone had ever thought possible. It was a true masterpiece, and a reminder of the importance of living in a way that is good for the \emph{environment}. The \emph{carriage} was a place where people could go to \emph{relax} and enjoy the peaceful atmosphere of the countryside, and it was a \emph{timely} reminder of the importance of preserving the past.

The \emph{carriage} was officially opened to the public again, and it quickly became a popular place to visit. People would come from all over to ride in the beautifully restored \emph{carriage}, enjoying the \emph{relaxing} atmosphere of the countryside. The \emph{carriage} also became a reminder of the importance of living in a way that is good for the \emph{environment}, and it was a place where people could go to learn about living \emph{environmentally}.

As the \emph{carriage}'s popularity grew, so did her reputation as a skilled worker and \emph{environmentalist}. She began to receive requests from all over the country to work on similar projects. She was happy to agree, using her skills to help preserve the past and promote \emph{environmentally} friendly living.

Years later, her work on the \emph{carriage} was recognized by a prestigious university, which offered her a full scholarship to study \emph{environmental} science. She was very happy and accepted the offer without hesitation. She went on to become a leading expert in her field, using her knowledge to help protect the \emph{environment} and preserve the past.

The old \emph{carriage} remained a beloved part of the town, a reminder of her hard work and dedication. It stood as a reminder of the power of imagination and perseverance, and the impact that one person can have on their community. And every time she sat in the \emph{carriage}, she was reminded of the importance of living in a way that is good for the \emph{environment}, and she was \emph{sure} that she had made a difference.

But what she did not expect was that the \emph{carriage} would also bring her a surprise. One day, as she was sitting in the \emph{carriage}, she met a man who would become her future husband. He was a historian, and he had come to the town to research the town's history. They struck up a conversation, and she showed him around the town, including the \emph{carriage}. As they sat together, they became close, and the \emph{carriage} became a special part of their love story, and it was a \emph{timely} and \emph{relaxing} experience for both of them."

\subsection{Polish}
An example of a story generated at B2 level for words \emph{dojeżdżać} (to commute), \emph{monitorować} (to monitor), \emph{mordować} (to murder), \emph{nieznaczny} (insignificant, minor), \emph{nowatorski} (innovative), \emph{ocaleć} (to be saved), \emph{piotrowy} (Peter’s), \emph{rejonowy} (district's), \emph{wynająć} (to rent, to hire), \emph{zwężać} (to narrow) is provided below.

W małym miasteczku, położonym w sercu Polski, mieszkał młody i pełen energii Piotr. Był on znanym i cenionym w swoim \emph{rejonie} za swoje \emph{nowatorskie} podejście do pracy. Piotr miał zwyczaj \emph{dojeżdżać} każdego dnia do pracy rowerem, co pozwalało mu na \emph{monitorowanie} sytuacji w mieście i jego okolicach.

Jednego dnia, Piotr otrzymał informację o tajemniczym \emph{morderstwie}, które miało miejsce w jednej z okolicznych wsi. Ofiarą był lokalny przedsiębiorca, znany ze złego postępowania. Piotr postanowił wyjaśnić sprawę i \emph{wynająć} prywatnego detektywa, aby pomógł mu w śledztwie.

W trakcie śledztwa, Piotr i detektyw odkryli, że \emph{morderstwo} nie było przypadkowe. Był to jeden z serii \emph{morderstw}, które miały miejsce w okolicy w ciągu ostatnich kilku tygodni. Sprawca, który okazał się być bardzo niebezpiecznym człowiekiem, \emph{mordował} swoje ofiary w okrutny sposób, pozostawiając za sobą tylko \emph{nieznaczne} ślady.

Piotr i detektyw postanowili \emph{monitorować} teren, aby znaleźć jakiekolwiek wskazówki, które mogłyby doprowadzić ich do sprawcy. W trakcie obserwacji, zauważyli, że w okolicy pojawia się coraz więcej patroli policyjnych, co mogło wskazywać na to, że policja również jest blisko rozwiązania sprawy.

Jednakże, Piotr i detektyw byli pewni, że są bliżej rozwiązania sprawy niż policja. Ich \emph{nowatorskie} podejście do śledztwa pozwoliło im na odkrycie kilku ważnych wskazówek, które mogły doprowadzić ich do sprawcy.

W końcu, po kilku dniach intensywnych poszukiwań, Piotr i detektyw znaleźli się w \emph{rejonie}, gdzie prawdopodobnie ukrywał się sprawca. Był to opuszczony budynek na obrzeżach miasta, który wyglądał jak idealne miejsce dla niebezpiecznego człowieka.

Gdy weszli do budynku, Piotr i detektyw zostali zaatakowani przez sprawcę. Był to moment, w którym musieli \emph{zwężać} jego pole widzenia, aby uniknąć ataku. Jednakże, dzięki swojej szybkości i szybkiej reakcji, udało im się \emph{ocaleć} i powstrzymać sprawcę.

Po aresztowaniu sprawcy, Piotr i detektyw zostali uznani za bohaterów. Ich \emph{nowatorskie} podejście do śledztwa i umiejętność \emph{monitorowania} sytuacji pozwoliły im na rozwiązanie sprawy, która wydawała się być niemożliwa do rozwiązania. Piotr powrócił do swojej pracy, ale teraz był znany nie tylko w swoim \emph{rejonie}, ale w całym kraju.

\subsection{Chinese}
An example of a story generated at HSK4 level for words: \begin{CJK*}{UTF8}{gbsn}单调 (monotonous), 团结 (to unite), 干什么 (what to do?), 必要 (necessary, indispensable), 打折 (discount), 把 (grammar particle), 旅客 (traveler), 条件 (condition, circumstance), 电影院 (cinema), 难道 (could it be?)\end{CJK*} is provided below.

\begin{CJK*}{UTF8}{gbsn}
李明觉得自己的生活很\underline{单调}，没有什么意思。每天都是重复的工作和生活，没有任何变化。他常常问自己：“我\underline{干什么}呢？\underline{难道}这就是我的生活吗？”

小张对李明说：“我们应该\underline{团结}起来，做一些有意义的事情。我们可以一起做志愿者，帮助那些需要帮助的人。”李明觉得这是一个好主意。他们开始一起做志愿者，帮助老人、病人和需要帮助的人。

他们还组织了一个社区活动，让所有的居民参加。活动中，他们有游戏、音乐和电影。李明觉得自己的生活突然变得有趣起来。小张说：“我们\underline{团结}起来，做一些有意义的事情，生活就会变得不再\underline{单调}。”

但是，有些人不理解他们的行为。他们觉得李明和小张是在浪费时间。他们说：“你们\underline{干什么}呢？这不是\underline{必要}的工作啊。”李明和小张回答说：“我们是在帮助别人，我们是在做有意义的事情。这是非常\underline{必要}的。”

有一天，李明和小张决定去\underline{电影院}看电影。他们买了票，准备进去的时候，发现\underline{电影院}正在\underline{打折}。他们觉得很幸运，能够以便宜的价格看电影。他们进去后，发现\underline{电影院}里有很多\underline{旅客}。他们觉得很奇怪，问\underline{电影院}的工作人员：“为什么这里有这么多\underline{旅客}？”工作人员回答说：“我们正在举办一个活动，很多人都来参加。”

李明和小张觉得很有趣。他们决定参加活动，和\underline{旅客们}一起看电影。他们发现，活动的\underline{条件}很好，电影也很精彩。他们觉得自己的生活变得更加丰富。小张说：“我们\underline{把}生活变得更加丰富了。”

李明突然想起了自己的问题：“\underline{难道}这就是我的生活吗？”他觉得自己已经找到了答案。他决定继续做志愿者，帮助别人，做有意义的事情。他觉得自己的生活已经变得不再\underline{	单调}。小张说：“我们\underline{团结}起来，生活就会变得更好。”
\end{CJK*}
\end{document}